\documentclass[11pt]{article}
\usepackage[T1]{fontenc}
\usepackage{amsmath,amssymb}
\usepackage{graphicx}
\usepackage{hyperref}
\usepackage[dvipsnames]{xcolor}
\newcommand{\rev}[1]{\textcolor{black}{#1}}
\usepackage{url}
\usepackage{amsthm}
\usepackage{algorithm, algorithmic}

\usepackage{titlesec}
\titlespacing*{\section}{0pt}{1ex}{1ex}
\titlespacing*{\subsection}{0pt}{0.6ex}{0.6ex}
\titlespacing*{\subsubsection}{0pt}{0.4ex}{0.4ex}

\titleformat{\section}
{\normalfont\fontsize{11}{11}\bfseries}{\thesection}{2em}{}
\titleformat{\subsection}
{\normalfont\fontsize{11}{11}\bfseries}{\thesubsection}{2em}{}

\titleformat{\subsubsection}
{\normalfont\fontsize{11}{11}\bfseries}{\thesubsubsection}{1em}{}

\definecolor{bittersweet}{rgb}{1.0, 0.44, 0.37}
\usepackage[most,skins,theorems]{tcolorbox}
\newtcolorbox{nbox}[1]{colframe=black,fonttitle=\bfseries, title=#1}

\usepackage[top=1in, bottom=1in, left=1in, right=1in]{geometry}
\title{Active Inference AI Systems for Scientific Discovery}
\author{ Karthik Duraisamy \\ {\em Michigan Institute for Computational Discovery \& Engineering,} 
	\\ {\em University of Michigan, Ann Arbor.}}
\date{}

\begin{document}

% ============================================================================
% REVISION NOTE: Text marked in \rev{red} indicates revisions made in response 
% to reviewer comments. The additional_refs.bib file contains new bibliography 
% entries required for the revisions.
% ============================================================================

\maketitle

\begin{abstract}
	The rapid evolution of artificial intelligence has led to expectations of transformative scientific discovery, yet current systems remain fundamentally limited by their operational architectures,  brittle reasoning mechanisms, and their separation from experimental reality. Building on earlier work--including foundational contributions in automated scientific discovery~\cite{langley2024integrated}, closed-loop laboratory systems~\cite{king2009automation,burger2020mobile,szymanski2023autonomous}, and causal machine learning~\cite{pearl2009causality,scholkopf2022causality}--this perspective contends that progress in AI-driven science now depends on closing three fundamental gaps—the abstraction gap, the reasoning gap, and the reality gap—rather than on model size/data/test time compute. While prior systems have made substantial progress on individual gaps, we argue that genuinely novel scientific discovery --defined here as the identification of previously unknown causal mechanisms, physical laws, or theoretical frameworks that generalize beyond training distributions-- requires their integrated resolution. Scientific reasoning demands internal representations that support  simulation of actions and response, causal structures that distinguish correlation from mechanism, and continuous calibration.
	Active inference AI systems for scientific discovery are defined as those that (i) maintain long-lived research memories grounded in causal self-supervised foundation models, (ii) employ symbolic or neuro-symbolic planners equipped with Bayesian guardrails, (iii) grow persistent knowledge graphs where thinking generates novel conceptual nodes, reasoning establishes causal edges, and real-world interaction prunes false connections while strengthening verified pathways, and (iv) refine their internal representations through closed-loop interaction with both high-fidelity simulators and automated laboratories—an operational loop where mental simulation guides action and empirical surprise reshapes understanding. In essence, this work outlines \rev{design principles for} an architecture in which discovery arises from the  interplay between internal models that enable counterfactual reasoning and external validation that grounds hypotheses in reality. It is also argued that the inherent ambiguity in  feedback from simulations and experiments, and  underlying  uncertainties makes human judgment indispensable, not as a temporary scaffold but as a permanent architectural component. 
\end{abstract}

%The trajectory of general AI capabilities—from transformers to reasoning models—parallels a quieter revolution in scientific applications. The pursuit of artificial general intelligence has produced systems capable of remarkable breadth but limited scientific depth. In contrast, transformative advances in specific domains—AlphaFold's solution to protein structure prediction and FourCastNet's revolution in numerical weather forecasting—represent "AlexNet moments" for their respective fields: narrow but profound breakthroughs. Yet a fundamental disconnect persists: general reasoning capabilities rarely translate to domain-specific discoveries, and specialized scientific models struggle with the creative leaps that define breakthrough science.

\section{Present day AI Systems and Scientific Discovery}
Over the past decade, the evolution of AI foundation model research has followed a clear sequence of discrete jumps in capability. The advent of the Transformer\cite{vaswani2017attention} marked a phase dominated by architectural innovations, which was rapidly succeeded by scaling demonstrations such as GPT-2\cite{radford2019language}. The maturation of large-language-model pre-training then gave way to the \rev{``usability turn'': the shift from models optimized purely for benchmark performance to chat-oriented systems} fine-tuned for alignment and safety that enabled direct human interaction\cite{ouyang2022training}. The current frontier is characterised by reasoning-emulation systems that incorporate tool use, scratch-pad planning, or program-synthesis objectives\cite{nye2021show}. A fifth, still-incipient phase points toward autonomous \rev{agents which are software systems capable of perceiving their environment, making decisions, and taking actions to achieve specified goals without continuous human supervision~\cite{wooldridge1995intelligent}}that can decompose tasks, invoke external software or laboratories, and learn from the resulting feedback. Scientific applications of AI have echoed each of these transitions at a compressed cadence.  As examples, SchNet translated architectural advances to quantum chemistry\cite{schutt2017schnet}; AlphaFold \rev{combined evolutionary multiple-sequence-alignment features with SE(3)-equivariant neural architectures~\cite{jumper2021highly} leveraging decades of accumulated protein structure data to achieve high accuracy, though still requiring petascale compute and vast unlabeled evolutionary sequences rather than dramatically fewer examples;} ChemBERTa~\cite{chithrananda2020chemberta} and FourCastNet~\cite{pathak2022fourcastnet} adapted language and vision innovations to molecular and climate domains; and AlphaGeometry applied reasoning-centric objectives to symbolic mathematics\cite{trinh2024solving}. Collectively, recent works~\cite{gridach2025agentic,biever2025ai,bran2024augmenting} chart a shift from single, specialized pre-trained model  to workflow orchestration, suggesting that future breakthroughs may hinge on integrating heterogeneous, domain-aware agents capable of planning experiments, steering simulations, and iteratively refining hypotheses across scales.

This highlights a deeper challenge for scientific discovery, which must reason across stacked layers of abstraction: the emergence of unexpected phenomena at higher scales, just as local atmospheric equations do not directly predict large-scale El Niño patterns\rev{~\cite{chalupka2015visual,chalupka2016unsupervised}}. \rev{This challenge of multi-scale abstraction has been recognized and discussed extensively in the automated scientific discovery literature~\cite{langley2024integrated}.} To address this challenge, it may be required  to deliberately architect systems with built-in mechanisms for hierarchical inference, equipping them with specialized components that can navigate between reductionist details and emergent phenomena. A compelling counter-argument posits that such abstract reasoning is not a feature to be explicitly engineered, but an emergent property that will arise from sufficient scale and diverse data. Proponents of this view might point to tools such as AlphaGeometry~\cite{trinh2024solving}, where complex, formal reasoning appears to emerge from a foundation model trained on vast synthetic data. However, we contend that while scaling can master any pattern present in the training distribution—even highly complex ones—it is fundamentally limited to learning correlational structures. Scientific discovery, in contrast, hinges on understanding interventional and counterfactual logic\rev{~\cite{pearl2009causality,hernan2020causal}}: {\em what happens when the system is deliberately perturbed?} This knowledge cannot be passively observed in static data; it must be actively acquired through interaction with the world or a reliable causal model thereof\rev{though we note that causal assumptions can sometimes be justified from observational data when appropriate structural constraints hold~\cite{cartwright2004causation,bareinboim2016causal}}. The `reality gap' thus remains a significant barrier that pure scaling may not cross.

%This  highlights a deeper challenge for scientific discovery, which must reason across stacked layers of abstraction: the emergence of unexpected phenomena at higher scales, just as local atmospheric equations do not directly predict large-scale El Niño patterns. %Can AI-driven computational systems recognize and create new levels of abstraction, or does this require something beyond pattern matching in high-dimensional spaces?

It is also pertinent to examine the nature of present-day scientific discovery before speculating the role of AI. Modern science has moved beyond the romanticized vision of solitary geniuses grappling with nature's mysteries. It may be difficult to generalize or even define the nature of discovery, but it is safe to assume that many of today's discoveries emerge from vast collaborations parsing petabytes of data from instruments such as the Large Hadron Collider or from distributed sensor networks or large-scale computations and most importantly, refining hypothesis in concurrence with experiments and simulations.  In fields such as high-energy physics, the bottleneck has shifted toward complexity management, whereas in data-constrained arenas such as fusion-plasma diagnostics, insight scarcity remains dominant; any general framework must therefore account for both regimes. Even if one possesses the raw data to answer profound questions, we often lack the cognitive architecture to navigate the combinatorial explosion of hypotheses, interactions, and emergent phenomena. \rev{The connection between cognitive architecture and the capacity for rich abstraction is direct: effective reasoning over scientific concepts requires working memory structures that can hold, manipulate, and compose abstract representations~\cite{lake2017building}, capabilities that current transformer architectures approximate but do not fully realize.} This creates an opportunity for AI systems—to excel precisely where human cognition fails, in maintaining consistency across very high-dimensional parameter spaces, identifying and reasoning about subtle patterns in noisy data.  At this juncture, it has to be emphasized that generating novel hypotheses might be the easy part~\cite{gu2024interesting}: the challenge is in rapidly assessing the impact of a hypothesis or action in an imaginary space. Thus AI systems have to be equipped with rich world models that can rapidly explore vast hypothesis spaces, and integrated with efficient computations and experiments to provide valuable feedback.

\rev{It is essential to distinguish between \textit{engineering discovery} -the optimization of known systems toward well-defined objectives- and \textit{scientific discovery} -the identification of novel causal mechanisms, physical laws, or theoretical frameworks. Engineering problems in materials science or drug design can often be addressed by closed-loop optimization systems operating over predefined search spaces~\cite{burger2020mobile,szymanski2023autonomous}. Scientific discovery, by contrast, requires the capacity for causal reasoning and theoretical insight that may restructure the problem space itself. Prior closed-loop systems have demonstrated genuine discoveries: Adam, an automated scientist for yeast metabolism, combined ontological reasoning with active learning to design and execute falsifiable experiments, leading to novel findings in genomics~\cite{king2009automation}; its successor Eve identified that triclosan is effective against malaria~\cite{williams2015cheaper}. These systems embody the active inference principles we advocate: maintaining scientific memory while engaging in closed-loop interaction with physical laboratories. The contribution of this perspective is not to claim that such systems are impossible, but to argue that scaling beyond well-defined domains requires addressing the abstraction, reasoning, and reality gaps in an integrated manner, and to propose architectural principles for doing so.}

%Yet this same complexity erects new barriers: scientific truth now requires consensus across heterogeneous evidence types, from genomic sequences to satellite imagery to social media traces, demanding AI systems that can reason across radically different epistemological frameworks while maintaining semantic coherence.  

Modern science also operates within a myriad of constraints that exist in economic, legal and social dimensions rather than physical laws. These constraints favor certain types of AI-driven discovery while effectively prohibiting others. Additionally, the  structure of the modern scientific enterprise—with its emphasis on incremental, publishable units and citation metrics—may be fundamentally misaligned with the kind of patient, integrative thinking required for paradigm shifts. AI systems could theoretically ignore these social pressures, pursuing research programs too risky or long-term for humans. But this same freedom from social constraint raises the possibility of AI systems optimizing for discovery without broader goals accounted for.  Perhaps more concerning, AI systems trained on existing scientific literature risk amplifying current biases and narrowing the space of explored ideas. In other words, AI systems might converge on well-studied paths~\cite{campo2025artificial,mateos2023there}, reducing the rich variety of research directions to a handful of statistically probable avenues. Scientific progress demands not convergence but divergence—an explosion of hypotheses, methodologies, and frameworks that challenge orthodoxy. The challenge is in designing AI systems that expand rather than constrain the landscape of scientific imagination.

Against this backdrop, the remainder of this perspective piece is organized around three interlocking hurdles that scientific discovery architectures must clear: (i) the abstraction gap, which separates low-level statistical regularities from the mechanistic concepts on which scientists actually reason; (ii) the reasoning gap, which limits today’s models to correlation-driven pattern completion rather than causal, counterfactual inference; and (iii) the reality gap, which isolates computation from the empirical feedback loops that ultimately arbitrate truth. Each gap both constrains and amplifies the others: without rich abstractions there is little substrate for reasoning, and without tight coupling to reality even the most elegant abstractions may drift toward irrelevance. \rev{While we treat these gaps in separate sections for expository clarity, they are deeply interconnected: the abstraction gap concerns \textit{what} representations the system can manipulate, the reasoning gap concerns \textit{how} those representations are transformed, and the reality gap concerns \textit{whether} those transformations correspond to the external world. Closing one gap in isolation yields limited progress; for instance, rich abstractions without causal reasoning remain correlational, while causal reasoning without empirical grounding may drift into unfalsifiable speculation.}

\begin{nbox}{Philosophical Foundations}
	The quest for  
	 scientific discovery via computation confronts a fundamental paradox.
	 Gödel~\cite{godel1931} proved that formal systems are incomplete, and cannot self-consistently prove all truths, while Wolfram~\cite{wolfram2002} demonstrates that computational irreducibility pervades nature. Additionally, Penrose~\cite{penrose1989,penrose1994} contends that human insight transcends algorithms. 
	%  Gödel~\cite{godel1931}, Wolfram~\cite{wolfram2002}, and Penrose~\cite{penrose1989,penrose1994} argue that formal systems cannot self-consistently prove all truths, computational irreducibility pervades nature, and human insight transcends  algorithms. 
	Yet scientific theories and computations are found to be highly effective in many cases. Insight can be gained from Wolfram's recent comment~\cite{wolfram2024ai}: {\em The very presence of computational irreducibility necessarily implies that there must be pockets of computational reducibility, where at least certain things are regular and predictable.  It is within these pockets of reducibility that science fundamentally lives.} 
	\vspace{0.15cm}
	
	\hspace{0.1cm} In most cases, these pockets cannot be deduced a priori—they require empirical discovery. This connects to Popper's~\cite{popper1959} falsificationism: we cannot prove we have found true reducibility, but we can discover boundaries through experiments that challenge assumptions. Empirical feedback escapes Gödel's constraints while delineating where nature permits shortcuts.
	Kuhn's analysis~\cite{kuhn1962} adds temporal dynamics: Science alternates between normal science within established pockets and paradigm shifts that restructure understanding.   AI systems must balance exploiting known regularities with flexibility to reconceptualize when evidence demands.
	\vspace{0.15cm}
	
	\hspace{0.075cm} 
	This synthesis directly informs our architecture. Thinking explores for new pockets and tests boundaries; reasoning exploits discovered regularities. World models encode provisional maps of known pockets, subject to Popper's falsification and Kuhn's paradigm shifts. 
	Human steering proves essential. Humans provide non-computational insight for recognizing genuine understanding, value judgments for directing exploration, and navigation through paradigm shifts where evaluation criteria themselves transform.  Humans can shape the search process by encoding domain knowledge, identifying significant anomalies, and recognizing connections that form larger frameworks. When Faraday discovered electromagnetic induction, he did not deduce it from Maxwell's equations (which did not yet exist)—he found it through experiment.
	Thus, productive collaborations can implement the complete scientific method: AI generates and tests hypotheses at scale; humans provide insight and judgment and empirical feedback provides critical steering.
\end{nbox}

\section{The Abstraction Gap}

While early models largely manipulated tokens and pixels, recent advances in concept‐bottleneck networks\cite{koh2020concept}, symmetry‐equivariant graph models\cite{thomas2018tensor}, and neuro-symbolic hybrids\cite{maoneuro} show preliminary evidence that contemporary AI can already represent and reason over higher‐order scientific concepts and principles. Yet a physicist reasons in conservation laws and symmetry breaking, whereas language models still operate on surface statistics. Closing this abstraction gap requires addressing several  intertwined weaknesses.

%\paragraph{Brittle Reasoning Architecture}
Modern transformer variants assemble chain-of-thought proofs\cite{wei2022chain} by replaying patterns observed during pre-training; they do not build explicit causal graphs or exploit formal logic engines except in narrow plug-in pipelines. As a result they fail at problems that demand deep compositionality. Several other shortcomings have also been pointed out~\cite{messeri2024artificial}\rev{: (i) AI systems can create ``illusions of understanding'' where researchers mistake fluent outputs for genuine insight; (ii) they may narrow the diversity of research directions by channeling attention toward well-represented topics in training data; and (iii) they can erode researchers' own reasoning skills through over-reliance on automated suggestions}.%True abstraction may require mechanisms for dynamic variable binding\cite{gershman2015structured}, iterative hypothesis testing, and analogy formation across ontology boundaries, perhaps by marrying differentiable programs\cite{mao2020differentiable} with automated theorem provers and symbolic simulators\cite{wu2024alphageometry}.

\begin{nbox}{Thinking and reasoning}

A critical distinction emerges between thinking and reasoning: Thinking can be operationalized as an iterative, exploratory process—searching for partial solutions in the form of patterns without guaranteed convergence. It is the slow, generative phase where new connections form and novel patterns emerge from a number of possibilities. Reasoning, by contrast, represents the fast, deterministic traversal of established knowledge structures—building the most expressive path through a graph of already-discovered patterns.
This dichotomy~\cite{johnson1983mental}  may explain why current AI systems excel at certain tasks while failing at others. Large language models can reason impressively when the requisite patterns exist in their training data—they rapidly traverse their learned knowledge graphs to construct seemingly intelligent responses. Yet they struggle with genuine thinking: the patient, iterative discovery of patterns that do not yet exist in their representational space. Scientific discovery demands both capabilities in careful balance. 
\vspace{0.15cm}

\hspace{0.1cm} 
Thinking generates hypotheses by discovering new patterns through mental simulation and exploration; reasoning then rapidly tests these patterns against existing knowledge and empirical constraints.
The purpose of thinking therefore is not to solve problems directly but to expand the pattern vocabulary available for subsequent reasoning. Each thinking cycle potentially adds new nodes and edges to the knowledge graph, creating shortcuts and abstractions that make previously intractable reasoning paths suddenly accessible. This is perhaps why breakthrough discoveries often seem obvious in retrospect—the thinking phase has restructured the problem space so thoroughly that the reasoning path becomes trivial. \rev{We note that thiw terminology differs from Kahneman's~\cite{kahneman2011thinking}  System 1/System 2 framework, where System 2 denotes slow, deliberate reasoning. Our usage follows Johnson-Laird's~\cite{johnson1983mental} mental models tradition, where ``thinking'' refers to the creative construction of new mental models (slow, exploratory), while ``reasoning'' refers to drawing inferences within established models (fast once patterns exist). The key insight is that scientific discovery requires both: the slow construction of novel conceptual frameworks \textit{and} the rapid verification of their logical consequences. Both frameworks recognize the importance of dual cognitive modes; the difference lies in which dimension (effort vs.\ novelty) defines the slow/fast distinction. Recent neuro-symbolic RL agents hint at this synergy:}
the survey of Acharya et al.\cite{acharya2023neurosymbolic} chronicles agents that fuse neural perception with first-order symbolic planners 
while Mao et al.\cite{maoneuro} demonstrate compositional question-answering by training a neural concept learner that hands off logic programs to a symbolic executor.
\end{nbox}

%\paragraph{Causal Reasoning as Mental Experimentation}

The gap between correlation and causation represents perhaps the most fundamental challenge in automated scientific discovery. While current models excel at finding statistical regularities, scientific understanding requires the ability to reason about interventions—to ask not just ``what correlates with what?" but ``what happens when we change this?"

Pearl's causal hierarchy~\cite{pearl2009causality} distinguishes three levels of cognitive ability: association (seeing), intervention (doing), and counterfactuals (imagining). Current AI systems operate primarily at the associative level, occasionally reaching intervention through experimental design. True scientific reasoning requires all three, particularly the counterfactual ability to imagine alternative scenarios that violate observed correlations. \rev{This claim finds support in the history of science: the link between smoking and cancer was established through epidemiological reasoning about counterfactuals without randomized controlled trials~\cite{cornfield1959smoking}. Similarly, Darwin's theory of evolution was fundamentally a counterfactual framework: imagining what would happen to populations under different selection pressures.} This connects directly to ethologist Konrad Lorenz's insight\rev{~\cite{lorenz1973behind}} --first tied to  learning systems by Scholkopf~\cite{scholkopf2022causality}\rev{, building on Pearl's foundational work~\cite{pearl2009causality}}-- that  thinking  is fundamentally about acting in imaginary spaces where we can violate the constraints of observed data. This {\em mental experimentation}—impossible in physical reality but accessible in the imagination—forms the basis of scientific law formation.

\section{Unhobbling Intelligence}

A certain level of consensus appears to be forming in the community that incremental scaling of present architectures may not deliver the qualitative leap that scientific discovery demands. Progress hinges on unhobbling—removing the design constraints that keep today’s models predictable, yet fundamentally limited—through concurrent advances in algorithms, speculation control, hardware co-design, and access models.

\paragraph{Algorithmic Gains} 

Future systems must balance the complementary modes of thinking and reasoning as first-class architectural principles. Thinking—or slow, iterative discovery of new patterns—demands (i) world-model agents that can explore counterfactual spaces through mental simulation  \cite{ha2018worldmodels}; (ii) curiosity-driven mechanisms that reward pattern novelty over immediate task performance\rev{~\cite{schmidhuber2010formal,pathak2017curiosity}}; and (iii) \rev{mechanisms for preventing premature convergence, such as entropy regularization~\cite{haarnoja2018soft} or explicit exploration bonuses that maintain hypothesis diversity during search~\cite{bellemare2016unifying}}. Reasoning—the fast, deterministic traversal of pattern graphs—demands (i) efficient knowledge graph architectures with learned traversal policies\rev{~\cite{das2018go}}; (ii) neuro-symbolic stacks that maintain both continuous representations and discrete logical structures\cite{maoneuro}; and (iii) caching mechanisms that transform expensive thinking outcomes into rapid reasoning primitives. The interplay between these modes mirrors how scientists alternate between exploratory experimentation (thinking) and theoretical derivation (reasoning).

The notion that ``{\em thinking is acting in an imaginary space}"—as Konrad Lorenz observed\rev{~\cite{lorenz1973behind}}—provides a foundational principle for understanding how world models enable scientific discovery. Just as biological organisms evolved the capacity to simulate actions internally before committing physical resources, AI systems with rich world models can explore vast hypothesis spaces through mental simulation. This capability \rev{enables a qualitatively different mode of inference compared to pure pattern matching}: it \rev{supports} counterfactual reasoning, experimental design optimization, and the anticipation of empirical surprises before they manifest in costly real-world experiments. 
World models can serve as the substrate for this imaginary action space, encoding not just correlations but causal structures that permit intervention and manipulation.  The fidelity of these mental simulations—their alignment with physical reality—determines whether the system's thoughts translate into valid discoveries.

Scientific progress thrives on disciplined risk: venturing beyond received wisdom while remaining falsifiable. Current alignment protocols deliberately dampen exploratory behaviour, biasing models toward safe completion of well-trodden trajectories. Controlled speculation frameworks—for example, curiosity-driven reinforcement learning~\cite{oudeyer2007intrinsic} combined with Bayesian epistemic guards—could allow systems to seek novel hypotheses, flag them with calibrated uncertainty, and propose targeted experiments for arbitration. Mechanisms such as self-consistency voting~\cite{wang2022selfconsistency}, adversarial peer review, and tool-augmented chain-of-thought audits offer additional scaffolding to keep high-variance reasoning tethered to empirical reality.

Recent empirical work by Buehler~\cite{buehler2025situ,buehler2025agentic} demonstrates that graph-based knowledge representations can bridge the abstraction gap. Specifically, recursive graph expansion experiments show that autonomous systems naturally develop hierarchical, scale-free networks mirroring human scientific knowledge structures. Without predefined ontologies, these systems spontaneously form conceptual hubs and persistent bridge nodes, maintaining both local coherence and global integration—addressing precisely the limitations that prevent current AI from connecting low-level patterns to high-level scientific concepts. Indeed, success in one class of problems does not guarantee translation to other problems, domains and disciplines, but these works show that with appropriate graph-based representations, AI systems can discover novel conceptual relationships.

\paragraph{Computational Inefficiency}
Scaling laws  show that models get predictably better with more data, parameter count and test time compute, yet every small gain  might come at a great expense in time and/or energy. Such brute-force optimization contrasts sharply with biological economies in which sparse, event-based spikes\cite{davies2018loihi} and structural plasticity\cite{kasthuri2015saturated} deliver continual learning at milliwatt scales. Bridging the gap will demand both algorithmic frugality—latent-variable models, active-learning curricula, reversible training—and hardware co-design. State-of-the-art foundation models require months of GPU time
and $> 10^{25}$ FLOPs to reach acceptable performance on long-horizon benchmarks.
Memory-reversible Transformers \cite{mangalam2022reversible,zhang2024exact} and curriculum training \cite{wanglearning} have recently reduced end-to-end \emph{training} costs by 30–45\,\%, without loss of final accuracy. Similar level of cost reductions have been reported~\cite{chung2024reducing} leveraging energy and power draw scheduling.

%\paragraph{Hardware–Algorithm Co-evolution and guardrails}
The von Neumann bottleneck—shuttling tensors between distant memory and compute—now dominates energy budgets~\cite{ibm_vonneumann_bottleneck_2024}. Processing-in-memory fabrics~\cite{zhang2024pim}, spiking neuromorphic cores that exploit event sparsity , analog photonic accelerators for low-latency matrix products, quantum samplers for combinatorial sub-routines~\cite{arute2019quantum}  could  open presently unreachable algorithmic spaces. Realising their potential outside of niche applications, however, will require co-design of hardware, software and algorithms and extensive community effort.
%\paragraph{Democratisation Without Dilution.}
%As foundation-model toolkits proliferate, the community must also balance broad access with methodological rigor. Concretely: open checkpoints should ship with provenance logs, uncertainty certificates, and reproducible evaluation metrics; compute-limited labs could receive federated fine-tuning grants that leverage shared base models while preserving data sovereignty; and journal review pipelines might require transparent model cards and code for experiment automation. Such guardrails ensure that widening participation accelerates discovery rather than amplifying error and hype.

\paragraph{Evaluations}
Current leaderboards—e.g. MathBench\cite{liu2023mathbench}, ARC\cite{chollet2019measure}, GSM8K\cite{cobbe2021training}—scarcely probe the generative and self-corrective behaviours central to science. A rigorous suite should test whether a model can (i) identify when empirical data violate its latent assumptions, (ii) propose falsifiable hypotheses with quantified uncertainty, and (iii) adapt its internal representation after a failed prediction. Concretely, this may involve closed-loop benchmarks\cite{tobin2023closed} in which the system picks experiments from a simulated materials lab, updates a dynamical model, and is scored on discovery efficiency; or theorem-proving arenas where credit is given only for proofs accompanied by interpretable lemmas. Without such stress-tests, superficial gains risk being mistaken for conceptual breakthroughs. 
Future evaluations can also assess the human-AI-reality-discovery feedback loop itself.  Early exemplars such as DiscoveryWorld~\cite{jansen2024discoveryworld}, PARTNR~\cite{chang2024partnr} and SciHorizon~\cite{qin2025scihorizon} represent steps towards this direction.

%\section{The Incompleteness of Pure Computation}
\section{Architecture for the Era of Experience}
Empirical feedback complements formal reasoning by supplying information inaccessible to purely deductive systems, thereby expanding—rather than mechanically escaping—the set of testable scientific propositions. The interplay between formal systems and empirical validation creates a bootstrap mechanism that circumvents  incompleteness and irreducibility constraints. This suggests that AI systems for discovery must be fundamentally open—not just to new data, but to surprise from reality itself. Scientific history abounds with internally coherent theories that later failed empirical tests, underscoring the indispensability of continuous validation against data. Current AI systems excel at interpolation within their training distributions but struggle with the extrapolation that defines discovery.  This is exacerbated by the fact that
 many scientific domains are characterized by sparse, expensive data and imperfect simulators. Unlike language modeling where data is abundant, a single protein crystallography experiment might take months and cost thousands of dollars. Simulations help but introduce their own biases—the ``sim-to-real gap" that plagues robotics extends to all of computational science. Our architecture must therefore implement a hybrid loop: physics priors guide ML surrogates, which direct active experiments, which update our understanding in continuous iteration.

\begin{nbox}{Scientific Intelligence and the `Bitter Lesson'}
	The interplay between imagination and experimental validation creates a synergistic process at the heart of scientific discovery. Thinking provides the hypothesis generation engine, while empirical feedback provides the selection pressure that refines these mental models toward truth. These ideas have also been popularized by Yann LeCun~\cite{lecun2022path} in the context of general machine intelligence. True understanding requires recognizing which causal relationships remain invariant as we move between the imaginary space of simplified models and the  reality of full-scale experiments.  The divergence between mental simulation and empirical outcome provides the richest learning signal. 
	\vspace{0.15cm}
	
	\hspace{0.1cm}
	Rich Sutton’s `bitter lesson’~\cite{sutton2019bitter} observes that scalable, general methods often outperform handcrafted heuristics, yet recent domain-aware models like AlphaFold illustrate that judiciously chosen scientific priors can accelerate learning within compute-intensive regimes. AlphaFold’s use of SE(3)-equivariant networks and multiple-sequence-alignment priors enabled high accuracy with $\approx 170 000$ labeled structures, but success still relied on petascale compute and vast unlabeled evolutionary data rather than a dramatic reduction from trillions of examples.
	%Thus, while general computation provides the engine of discovery, it is the continuous grounding in empirical reality that steers this engine toward truth, preventing it from drifting into internally consistent but physically irrelevant speculation.
	\vspace{0.15cm}
	
	\hspace{0.1cm}
	We propose a `complete lesson': intelligent systems arise from computation plus constraints with feedback from reality. The `bitter lesson' correctly observes that general learning methods eventually outperform human-engineered heuristics within a fixed computational domain. However, it implicitly assumes that all relevant information is already encoded in the data. We contend that the most critical constraint for scientific intelligence is not a clever, human-designed inductive bias, but the information-rich feedback from the physical world, a qualitatively different signal that contains latent information about causality, invariance, and the validity of the model's core assumptions that may be absent from any tractably finite, static dataset.
	
	The key insight is knowing when to leverage existing knowledge versus when to remain agnostic: When exploring many engineering domains (drug design, materials science), encode known physics. When questioning the nature of reality (quantum gravity, consciousness), avoid premature constraints. It is worth noting that much of science and engineering  relies on the former scenario, and thus the art lies in choosing the right inductive biases for the problem at hand.  This does not represent a retreat from the bitter lesson but its fulfillment:  learning what to search for and how to search, guided by accumulated scientific knowledge and empirical feedback.
\end{nbox}

\paragraph{Causal  Models}

The current paradigm of domain-specific foundation models—from protein language models to molecular transformers—represents significant progress in encoding domain knowledge. However, these models fundamentally learn correlational patterns rather than causal mechanisms. ChemBERTa can predict molecular properties through pattern matching but cannot simulate how modifying a functional group alters reaction pathways. AlphaFold predicts protein structures through evolutionary patterns but does not model the physical folding process.

Scientific discovery demands  models that transcend pattern recognition to capture causal dynamics. A causal molecular  model would not just recognize that certain molecular structures correlate with properties—it would explain how electron density distributions cause reactivity,  and how thermodynamic gradients drive reactions. This causal understanding enables the counterfactual reasoning essential to science: predicting outcomes of novel interventions never seen in training data. This architectural choice has profound implications: foundation models scale with data and compute, but causal models scale with understanding. As we accumulate more structural data, foundation models improve at interpolation. As we refine causal mechanisms, foundation models improve at extrapolation—the essence of scientific discovery.

\paragraph{Physics priors}
While generative models like Sora create visually compelling outputs, they lack physical consistency—objects appear and disappear, gravity works intermittently, and causality is merely suggested rather than enforced. Mitchell~\cite{mitchell1980need} states that without biases to prefer some generalizations over others, a learning system cannot make the inductive leap necessary to classify instances beyond those it has already seen. Such inductive biases or physics priors—can be built-in to ensure generated realizations obey conservation laws, maintain object permanence, and support counterfactual reasoning about physical interactions.

Recent implementations demonstrate that world models can also discover physical laws through interaction. The joint embedding predictive architecture ~\cite{assran2023self,assran2025v} learns to predict object movements without labeled data, suggesting that the feedback loop between mental simulation and empirical observation can be implemented through self-supervised learning objectives that reward accurate forward prediction.
Current world models and coceptualizations thereof, however,  remain limited to relatively simple physical scenarios. While they excel at rigid body dynamics and basic occlusion reasoning, they are generally insufficient to describe complex phenomena like fluid dynamics or emergent collective behaviors. This gap between toy demonstrations and the full complexity of scientific phenomena represents the next frontier.

\begin{figure}
	\centering
	\includegraphics[width=0.8\textwidth]{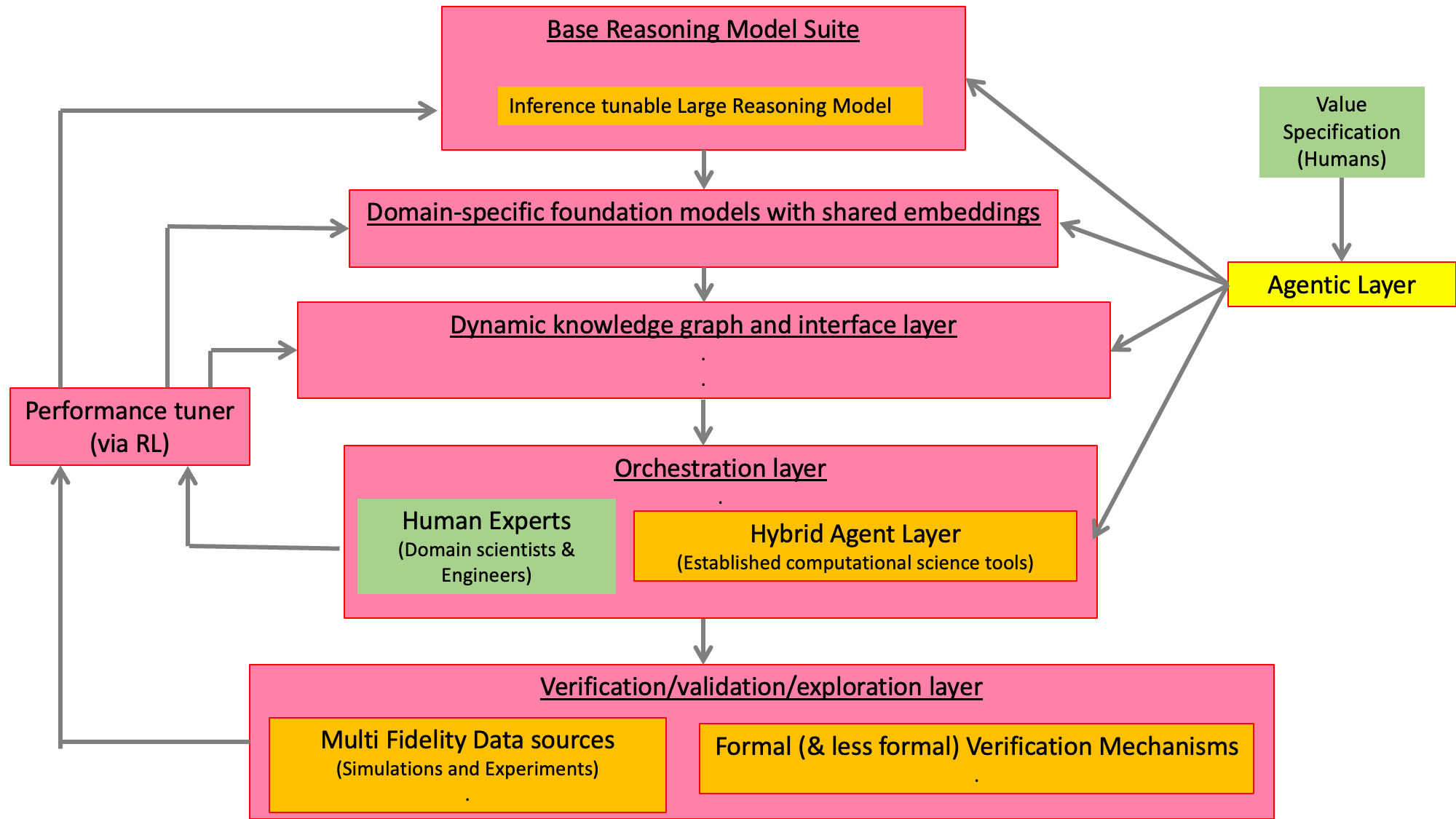}
	\caption{Exemplar architecture of an Active Inference AI system for scientific discovery. }
	\label{fig1}
\end{figure}

\paragraph{Active Inference AI Systems to Navigate Complex Scientific Questions} Many scientific phenomena exhibit chaotic dynamics, multiscale interactions, and emergent properties that defy reductionist analysis. Climate systems, biological networks, and turbulent flows operate across scales from molecular to planetary. Traditional ML approaches that assume smooth, well-behaved functions fail catastrophically in these domains. We need architectures that can reason across scales, identify emergent patterns, and know when deterministic prediction becomes impossible. No single formal or informal computational system can accomplish these tasks, and hence we propose an AI stack. An exemplar architecture is shown in Figure~\ref{fig1}. Some of the components of the architecture include:
\begin{enumerate}
	\item {\em Base reasoning model suite with inference-tunable capabilities:} This top-layer component comprises large reasoning models that can dynamically adjust their inference strategies based on the problem context. In contrast to  being optimized for next-token prediction, these models support extended thinking times, systematic exploration of solution paths, and explicit reasoning chains. The suite has the ability to recognize which mode of reasoning is appropriate. Value specifications from humans guide the reasoning process, ensuring that  resources are allocated to scientifically meaningful directions rather than arbitrary pattern completion.
\item {\em Multi-modal domain foundation models with shared representations:} These are effectively world models that maintain causal representations of scientific domains. These models allow the system to mentally simulate interventions, test counterfactuals, and explore hypothesis spaces before committing to physical experiments. These function as oracles or world models, serving as the substrate for both pattern discovery (thinking) and rapid inference (reasoning).
These domain-specific models must share embeddings that enable cross-pollination of insights. 
\item {\em Dynamic knowledge graphs as evolving scientific memory:} Unlike static knowledge bases, these graphs function as  cognitive architectures that grow through the interplay of thinking, reasoning, and experimentation. Nodes represent concepts ranging from raw observations to abstract principles, while weighted edges encode causal relationships with associated uncertainty. The graphs expand as thinking discovers new patterns (adding nodes), reasoning establishes logical connections (adding edges), and experiments validate or falsify relationships (adjusting weights). Version-controlled evolution allows the system to maintain competing hypotheses, track conceptual development, and recognize when anomalies demand fundamental restructuring rather than incremental updates. This persistent, growing memory enables genuine scientific progress rather than mere information retrieval.
\item {\em Reality tethering through verification layers:} The verification layer partitions scientific claims into formally provable statements and empirically testable hypotheses. Mathematical derivations, algorithmic properties, and logical arguments can be decomposed into proof obligations for interactive theorem provers (Lean~\cite{moura2021lean}, Coq~\cite{bertot2013interactive}), creating a growing corpus of machine-verified knowledge that future reasoning can build upon. For claims beyond formal correctness—predictions about physical phenomena, chemical reactions, or biological behaviors—the system generates targeted computational simulations and experimental protocols. This dual approach acknowledges that scientific knowledge spans from mathematical certainty to empirical contingency.  Crucially, failed verifications become learning opportunities, updating the system's confidence bounds and identifying gaps between its world model and reality. \rev{When formal and empirical verification diverge for instance, when a mathematically valid prediction fails experimentally, the system must flag this inconsistency for human review, as the divergence may indicate modeling assumptions that require revision rather than simple parameter updates.}
\item {\em Human-steerable orchestration:} Humans excel at recognizing meaningful patterns and making creative leaps; AI can perform exhaustive search and maintaining consistency across vast knowledge spaces; Well-understood computational science tools (e.g. optimal experimental design) can execute efficient agentic actions in a reliable manner.  This symbiotic relationship ensures that the system's powerful reasoning capabilities remain tethered to meaningful scientific questions, and existing algorithms are efficiently leveraged. \rev{We note that optimal experimental design methods can be inadequate when conditioned on misspecified models~\cite{sloman2022robustness}; human oversight is therefore essential not just for value alignment but for detecting when the system's foundational assumptions require revision.}
\item {\em Proactive exploration engines:} Rather than passively responding to queries (the primary mode in which language models are used currently), these systems work persistently in the backgrond to generate hypotheses, identify gaps in knowledge, and propose experiments. Driven by uncertainty quantification and novelty detection algorithms, these engines can maintain a priority queue of open questions ranked by their potential to achieve specified goals versus resource requirements. This layer enables the system to operate across multiple time horizons—pursuing rapid experiments vs long-term research campaigns that systematically map uncharted territories in the knowledge space. \rev{The ranking of scientific goals reflects value judgments that must ultimately derive from human priorities; the system provides estimates of expected information gain and resource costs, but the weighting between scientific impact, feasibility, and broader societal considerations remains a human responsibility. The tension between uncertainty-driven exploration and novelty-driven exploration remains an open challenge~\cite{dubova2022against}; current approaches typically require domain-specific tuning rather than universal solutions.}
\end{enumerate}

\rev{
\paragraph{Component Interactions}
Figure~\ref{fig1} illustrates the flow of information between components. The base reasoning model receives queries from human users and orchestrates calls to domain foundation models, which return predictions with uncertainty estimates. These predictions update the dynamic knowledge graph, which in turn informs the verification layer's decisions about whether claims require formal proof, computational simulation, or physical experimentation. The performance tuner (via reinforcement learning) adjusts exploration-exploitation tradeoffs based on reward signals derived from successful predictions, experimental confirmations, and human feedback on the utility of discoveries. Concretely: (i) reward signals for the RL tuner originate from prediction accuracy on held-out data, confirmation rates in downstream experiments, and explicit human ratings of hypothesis quality; (ii) the time horizons  are domain-dependent. The architectural principles outlined here are intended as design guidance rather than a complete implementation specification; realizing them in practice will require substantial engineering and domain-specific adaptation.
}

The architectural principles outlined above find grounding in recent work on transformational scientific creativity. For instance, Schapiro et al.~\cite{schapiro2025transformational}  formalize scientific conceptual spaces as directed acyclic graphs, where vertices represent generative rules and edges capture logical dependencies. This offers a concrete implementation pathway for the proposed dynamic knowledge graphs. Their distinction between modifying existing constraints versus fundamentally restructuring the space itself maps directly onto our architecture's dual modes of reasoning (traversing established knowledge) and thinking (discovering new patterns that may violate existing assumptions). This convergence suggests that achieving transformational scientific discovery through AI systems requires systems capable of identifying and modifying the foundational axioms that constrain current scientific understanding—a capability  the active inference framework aims to provide through its stacked architecture and integration of  models, empirical feedback, and human guidance.

It is acknowledged that while the AI system can, in principle, be operated  autonomously through well-defined interfaces between components, human interaction and decisions can be expected to play a key role. The architectural principles outlined above find partial instantiation in contemporary systems, though none fully realize the complete vision of  scientific intelligence.  Appendix A examines some current implementations  through the lens of our three-gap framework, and discusses both substantial progress and persistent limitations that illuminate the path forward. Appendix B gives high-level mathematical constructs for key components of the above system.

\paragraph{Challenges of Iterative Learning and Importance of Human Interactions}
While the aforementioned architecture presents a compelling vision of AI systems that learn from real-world interaction, incorporating  feedback into iterative training poses fundamental challenges that cannot be overlooked. Scientific experiments  produce sparse, noisy, and often contradictory signals. A single failed synthesis might stem from equipment miscalibration, modeling errors, or genuine chemical impossibility—yet the system must learn appropriately from each case. The tension between generalization and specificity becomes acute: overfitting to particular configurations may yield brittle models that fail to transfer across laboratories, while excessive generalization may miss critical context-dependent phenomena. 

This inherent ambiguity in processing experimental feedback into actionable model refinements makes human judgment indispensable, not as a temporary scaffold but as a permanent architectural component.  Thus, the challenge lies not merely in designing systems that can incorporate feedback, but in creating architectures that handle the full spectrum of empirical reality, including clear confirmations, ambiguous results,  systematic biases and truly novel results.  Effective human-AI collaboration must therefore go beyond simple oversight\rev{empirical studies demonstrate that human-AI teams can outperform either alone, but only when humans maintain genuine understanding rather than deferring uncritically to AI suggestions~\cite{bansal2021does}}. This partnership becomes especially critical when experiments and computations  challenge fundamental assumptions.

\section{Outlook}

%The deepest lesson from cognitive science and ethology is that intelligence—whether biological or artificial—fundamentally involves the capacity to act in imaginary spaces. %Scientific discovery amplifies this principle: it requires not just simulating within known physics but imagining new physics entirely. %The Lorenzian view of thinking as virtual action provides a unifying framework for understanding how world models, causal reasoning, and empirical grounding must work together in discovery systems.
Active AI systems encompass  external experience (empirical data) and internal experience (mental simulation). AI systems that can fluidly navigate between these modes will mark the transition from tools that find patterns to partners that discover principles. This perspective builds upon substantial progress in causal machine learning, active learning, and automated scientific discovery while addressing critical gaps. The causal machine learning community has made significant strides in developing methods for causal inference from observational data, with frameworks like Pearl's causal hierarchy and recent advances in causal representation learning providing mathematical foundations for understanding interventions and counterfactuals\rev{~\cite{scholkopf2021toward,ke2022learning}}. Similarly, active learning has evolved sophisticated strategies for optimal experimental design\rev{~\cite{settles2012active,gal2017deep}}, while automated discovery systems have demonstrated success in specific domains such as materials science and drug discovery\rev{~\cite{strieth2020machine,swanson2025virtual}}. However, these communities have largely operated in isolation, with causal methods focusing primarily on statistical inference rather than physical mechanism discovery, active learning optimizing for narrow uncertainty reduction rather than conceptual breakthrough, and automated discovery systems excelling at interpolation within known spaces rather than extrapolation to genuinely novel phenomena. \rev{We note that efforts to bridge these communities are underway: the Acceleration Consortium~\cite{accelerationconsortium2024} represents a major initiative integrating autonomous laboratories with AI-driven discovery, while recent work demonstrates promising integration of causal reasoning with materials optimization~\cite{hino2020active,pablo2019new,butler2018machine}.} 

Current implementations prioritize task completion over understanding, optimization over exploration, and correlation over causation. The path forward requires  AI systems that integrate causal reasoning not merely as a statistical tool but as the foundation for mental simulation and counterfactual experimentation, extending active learning beyond data efficiency to include the generation and testing of novel hypotheses that violate existing assumptions, and grounding automated discovery in continuous empirical feedback loops that prevent drift from physical reality. Most critically, while existing approaches excel within their prescribed domains, they lack the architectural foundation for the kind of open-ended, cross-domain reasoning that characterizes human scientific discovery—the ability to recognize when anomalous observations demand not just parameter updates but fundamental reconceptualization of the problem space itself.

Scientific discovery has always been a collaborative enterprise—across disciplines, institutions, and generations. AI systems represent new kinds of collaborative tools.  The transition from static models to living systems marks a fundamental shift in how we conceive of AI systems that persist, that remember, that build intuition through repeated engagement with reality. Just as human scientists develop insight through years of experimentation, future AI systems will accumulate wisdom through continuous cycles of hypothesis, experiment, and revision. We thus call for the creation of new benchmarks and research programs centered around the proposed stacked architecture, moving evaluation beyond static datasets to interactive, discovery-oriented environments. \rev{Success should ultimately be judged not only by benchmark performance but by domain expert assessment of discovery quality. For example, whether AI-generated hypotheses lead to experiments that domain scientists find surprising and informative, whether the system identifies gaps in current understanding that experts recognize as important, and whether it proposes experimental designs that are both feasible and scientifically meaningful. Structured evaluation protocols, such as blinded assessment of AI-proposed versus human-proposed hypotheses by independent expert panels, could provide more rigorous qualitative evaluation~\cite{musslick2025automating}.}

Finally, it has to be emphasized that modern AI systems are already useful in their present form, and are being utilized effectively by scientific research groups across the world. However, even with future  improvements, these tools  bring many systemic hazards~\cite{messeri2024artificial}: {\em a) false positives and false negatives:} spurious correlations can be mistaken for laws, while cautious priors may hide real effects, and thus rigorous uncertainty metrics and adversarial falsification must be built in; {\em b) epistemic overconfidence:} large models \rev{can exhibit poorly calibrated uncertainty estimates, particularly on inputs far from the training distribution, a phenomenon where predictive confidence remains high even when accuracy degrades~\cite{guo2017calibration}}; ensemble disagreement \rev{among multiple independently trained models} provides one diagnostic for such failures; {\em c) erosion of  insight and rigor:} over time, there is signficiant risk of researchers losing key scientific skills; {\em d) Cost:} simulation-driven exploration can consume resources long after \rev{the expected reduction in uncertainty from additional experiments becomes negligible relative to their cost (i.e., marginal information gain saturates)}; \rev{resource allocation algorithms (schedulers)} must weigh value against resources; {\em e) \rev{instrumental} drift:} \rev{scientific instruments and sensors evolve over time, and their characteristics may shift due to wear, recalibration, or environmental changes}; \rev{this can be addressed through standard calibration practices such as round-robin testing across laboratories~\cite{roundrobin2024}, but} without continual residual checks and rapid retraining, predictions may silently bias. These issues have to be continually acknowledged, recognizing and safeguards should be embeded into the scientific process.

%\rev{We acknowledge that this perspective is  speculative in nature, proposing design principles whose efficacy must ultimately be demonstrated through implementation and empirical evaluation. The probability of any particular architectural choice proving optimal is difficult to estimate; we offer these principles as a framework for organizing future research rather than as a definitive solution. The value of this perspective lies in identifying the key gaps that must be addressed and proposing concrete architectural responses, while recognizing that the specific implementations will require substantial iteration and domain-specific adaptation.}

\section*{Acknowledgment}
This piece has benefitted directly or indirectly from many discussions with Jason Pruet (OpenAI), Venkat Raman, Venkat Viswanathan, Alex Gorodetsky (U. Michigan), Rick Stevens (Argonne National Laboratory/U. Chicago), Earl Lawrence (Los Alamos National Laboratory) and Brian Spears (Lawrence Livermore National Laboratory). This work was partly supported by  Los Alamos National Laboratory under the grant  \#AWD026741 at the University of Michigan.

{
\small
\bibliographystyle{plain}
%\bibliographystyle{elsarticle-num}
% NOTE: The additional references required for revisions are in additional_refs.bib
% These should be merged with the existing refs.bib before compilation
\bibliography{refs,additional_refs}
}

\section*{Appendix A:  Current Implementations of Agentic Systems}
	A comprehensive review of agentic systems for scientific discovery can be found in Ref.~\cite{gridach2025agentic}\rev{; see also Zenil et al.~\cite{zenil2023future} for a complementary perspective}. Below, a few references that are related to abstraction, reasoning and reality gaps are provided.
	
 \rev{Before the transformer era, several  systems demonstrated that AI could make genuine scientific discoveries through closed-loop interaction with physical laboratories. Adam and Eve~\cite{king2009automation,williams2015cheaper} combined ontological reasoning with active learning to autonomously design and execute falsifiable experiments.  More recently, the Robot Chemist~\cite{burger2020mobile} demonstrated autonomous exploration of reaction conditions, while A-Lab~\cite{szymanski2023autonomous} achieved autonomous synthesis of novel materials. These systems embody key principles of our proposed architecture: persistent scientific memory (through ontologies and databases), closed-loop empirical validation, and hypothesis-driven experimentation. However, they operate within well-defined search spaces with predetermined objectives, limiting their capacity for the kind of open-ended discovery that might restructure the problem space itself.}
	
	Recent systems demonstrate varying degrees of success in elevating from statistical patterns to scientific abstractions. ChemCrow~\cite{bran2024augmenting} integrates eighteen expert-designed tools to bridge token-level operations with chemical reasoning, enabling tasks such as reaction prediction and molecular property analysis.  ProtAgents~\cite{ghafarollahi2024protagents} employs reinforcement learning to navigate the conceptual space of protein design, moving beyond sequence statistics to optimize for biochemical properties. Agent Laboratory's~\cite{schmidgall2025agent}  achieves high success rates in data preparation and experimentation phases while exhibiting notable failures  during literature review. 
	
	The reasoning gap manifests most clearly in  limited capacity for genuine causal inference. Coscientist~\cite{boiko2023autonomous} represents the current frontier, successfully designing and optimizing cross-coupling reactions through iterative experimentation, though its reasoning remains fundamentally correlational. LLaMP~\cite{chiang2024llamp} attempts to address this limitation by grounding material property predictions in atomistic simulations, effectively implementing a preliminary form of mental experimentation. These systems, while promising, cannot yet perform the counterfactual reasoning that distinguishes scientific understanding from mere pattern matching.
	
	The reality gap presents both tangible progress and stark limitations. Systems such as Organa~\cite{darvish2024organa} demonstrate sophisticated integration with laboratory robotics, automating complex experimental protocols in electrochemistry and materials characterization. CALMS~\cite{prince2024opportunities} extends this integration by providing context-aware assistance during experimental execution. However, these implementations reveal  brittleness: when experimental outcomes deviate from expected patterns, current systems lack the adaptive capacity to reformulate hypotheses or recognize when their fundamental assumptions require revision.

	Multi-agent architectures such as BioInformatics Agent~\cite{xin2024bioinformatics} and CellAgent~\cite{xiao2024cellagent} represent attempts to address these limitations through specialized collaboration, with distinct agents handling data retrieval, analysis, and validation. While these systems demonstrate improved performance on well-structured tasks, they do not yet perform the open-ended exploration that characterizes genuine discovery. The coordination overhead and brittleness of inter-agent communication often negate the benefits of specialization when confronting novel phenomena.
	
	These implementations and others are already accelerating science, but also collectively reveal a critical insight: current systems excel at automating well-defined scientific workflows but falter when required to navigate the uncertain terrain of genuine discovery. They can execute sophisticated experimental protocols, analyze complex datasets, and even generate plausible hypotheses, yet they lack the metacognitive capabilities to recognize when they are operating beyond their training domains. \rev{The contribution of this perspective is to identify the architectural requirements for moving beyond these limitations, not to claim that closed-loop discovery is impossible, but to argue that scaling to genuinely novel discovery requires integrated resolution of the abstraction, reasoning, and reality gaps.}

\end{document}